\documentclass[10pt,twocolumn,letterpaper]{article}

\usepackage{iccv}
\usepackage{times}
\usepackage{epsfig}
\usepackage{graphicx}
\usepackage{amsmath}
\usepackage{amssymb}

\usepackage{iccv}
\usepackage{times}
\usepackage{epsfig}
\usepackage{graphicx}
\usepackage{amsmath}
\usepackage{amssymb}
\usepackage{amsthm}
\usepackage{float}
\usepackage{mathtools}
\usepackage{tabularx}
\usepackage{multirow}
\usepackage{booktabs}
\usepackage{subfloat}
\usepackage{color}
\usepackage[normalem]{ulem}
\usepackage{bm}
\usepackage{calc}
\usepackage{subcaption}
\usepackage{enumitem}
\newtheorem{definition}{Definition}
\newtheorem{theorem}{Theorem}

\usepackage{algorithm} 
\usepackage{algpseudocode} 

\newcommand{\bfsection}[1]{\vspace*{0.1cm}\noindent\textbf{#1}}

\usepackage[breaklinks=true,bookmarks=false]{hyperref}

\iccvfinalcopy 

\def\iccvPaperID{****} 

\ificcvfinal\fi

\begin{document}

\title{Leveraging Superfluous Information in Contrastive Representation Learning}


\author{Xuechu Yu$^1$\ \ \
Fangzhou Lin$^{1}$\ \ \
Yun Yue$^1$\ \ \
Ziming Zhang$^{1}$\ \ \\
$^1$Worcester Polytechnic Institute, USA \\
\tt\small
\{xyu4, flin2, yyue, zzhang15\}@wpi.edu \\
\tt\small
}

\maketitle
\ificcvfinal\thispagestyle{empty}\fi

\begin{abstract}
   Contrastive representation learning, which aims to learn the shared information between different views of unlabeled data by maximizing the mutual information between them, has shown its powerful competence in self-supervised learning for downstream tasks. However, recent works have demonstrated that more estimated mutual information does not guarantee better performance in different downstream tasks. Such works inspire us to conjecture that the learned representations not only maintain task-relevant information from unlabeled data but also carry task-irrelevant information which is superfluous for downstream tasks, thus leading to performance degeneration. In this paper we show that superfluous information does exist during the conventional contrastive learning framework, and further design a new objective, namely {\em SuperInfo}, to learn robust representations by a linear combination of both predictive and superfluous information. Besides, we notice that it is feasible to tune the coefficients of introduced losses to discard task-irrelevant information, while keeping partial non-shared task-relevant information according to our {\em SuperInfo} loss.We demonstrate that learning with our loss can often outperform the traditional contrastive learning approaches on image classification, object detection and instance segmentation tasks with significant improvements.
\end{abstract}

\section{Introduction}

Due to the huge cost in acquiring data notations, unsupervised learning has enjoyed its renaissance recently. Contrastive learning, whose goal is to learn powerful presentations for downstream tasks, has achieved promising success \cite{wu2018unsupervised,devlin2018bert,misra2020self,tian2020contrastive,he2020momentum,grill2020bootstrap,gao2021simcse,radford2021learning}. Since there is no label information, contrastive learning usually estimates the mutual information between the learned representations of different views as its objective function such as SimCLR \cite{chen2020simple}, and takes the learned model as a features extractor for various downstream tasks, such as image classification,  object detection and instance segmentation.

\begin{figure}[htb]
\centering
\includegraphics[width=\columnwidth]
{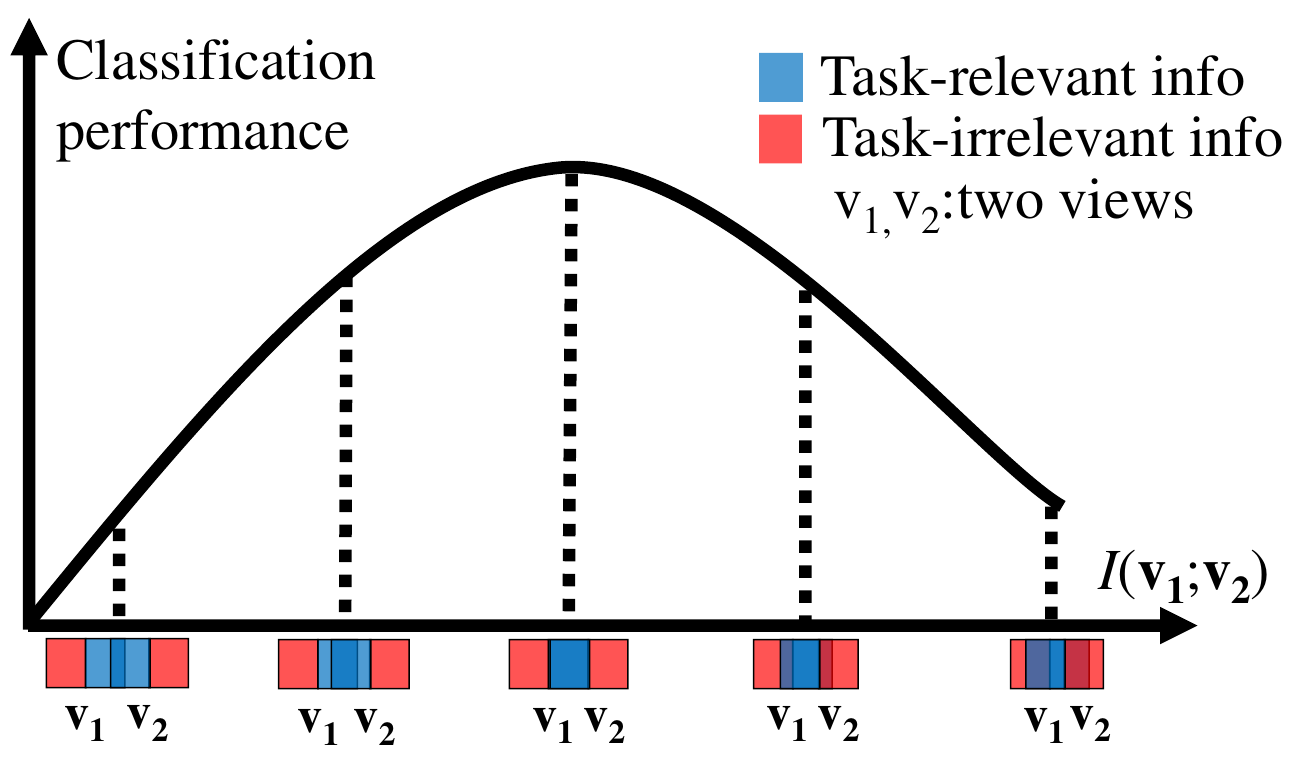}
\caption{Classification performance \vs estimated mutual information between the two views}
\label{fig1}
\end{figure}

To understand contrastive learning, we are often based on a multi-view assumption \cite{sridharan2008information,xu2013survey}: either view of the unlabeled data is (approximately) sufficient for the label information so that we can pull similar data pairs close while pushing dissimilar pairs apart to obtain the shared information of different views. From this perspective, contrastive learning aims to maximize the lower bound of the mutual information between two augmentation views of the data so that the learned presentations are useful for downstream tasks. However, some researchers find counterintuitive examples. For instance, \cite{tschannen2019mutual,tian2020makes} argue that maximizing the lower bound of the mutual information of the learned representations can not guarantee good performance for various downstream tasks, there are other factors which can not be ignored. \cite{tian2020makes} did a series of experiments that reveals the relationship between the estimated mutual information between two augmentation views and downstream image classification performance (performance on CIFAR-10 \cite{krizhevsky2009learning} and STL-10 \cite{coates2011analysis}), as shown in Figure \ref{fig1}, with different data augmentations, the estimated mutual information between two augmentation views becomes larger and larger, however, the downstream classification performance does not vary with the changed estimated mutual information, but rather goes up at first and then goes down.

Based on this phenomenon, we guess that the learned representations not only extract task-relevant information from the input data, but also carry task-irrelevant information which is superfluous for various downstream tasks, thus leading to performance degeneration. For the supervised learning, prior work \cite{alemi2016deep} straightforwardly discards task-irrelevant information from the input data by maximizing the mutual information between the learned representation and the label, while simultaneously minimizing the mutual information between the learned representation and the input data to make the learned representation more sufficient. While in the self-supervised framework, since there is no provided label information, each view plays a supervisory role for the other one, what we pay attention to is the shared information between different augmentation views. Through detailed analysis analogously to supervised learning, 
we find that the mutual information between each augmentation view and its encoding is comprised of two components, task-relevant one and task-irrelevant one, further we can express each part with the notation of mutual information between different variables. As a consequence, we create a new objective function to remove task-irrelevant information while maximizing the mutual information between two different augmentation views, which can improve performance for various downstream tasks. Besides, we notice that we can tune the coefficients of introduced losses to discard task-irrelevant information, while simultaneously keeping partial non-shared task-relevant information according to different tasks. What's more, we draw a conclusion that the learned representations from our method can have a better performance than others on the downstream tasks when the multi-view redundancy is small by analyzing the Bayes Error Rate of different representations (See Section \ref{Bayes}).

Overall, our contributions include:
\begin{itemize}[nosep, leftmargin=*]
\item Based on prior works and supervised learning solutions, we excavate two independent parts, the task-relevant part and the task-irrelevant part, that make up the mutual information between each augmentation view and its encoding, and express each part with mutual information in different variables. Consequently, we design a new objective function to eliminate the task-irrelevant part.

\item By applying the theory of Bayes Error Rate, we prove that our learned presentations can perform better on the downstream tasks.

\item We verify the effectiveness of our method for contrastive representation learning framework by conducting image classification, object detection and instance segmentation experiments. What's more, we also run certain ablation experiments to analyze the role of the adding losses.
\end{itemize}

\section{Related Work}

Contrastive representation learning, one of the several self-supervised learning approaches, has significantly outperformed other approaches from recent years \cite{soatto2014visual,zhang2017split,kingma2013auto,pathak2016context,donahue2019large,noroozi2016unsupervised,doersch2015unsupervised,gidaris2018unsupervised,zhang2019aet,yue2023hyperbolic,chen2021exploring,zbontar2021barlow}. With the convenience of obtaining a mass of unlabeled data, different multi-views of unlabeled data are constructed to design the specific contrastive loss to obtain a powerful learned representation, such as multiple augmentations of one image \cite{bachman2019learning,ye2019unsupervised,srinivas2020curl,zhao2021distilling,zhuang2019local}, different patches of one image \cite{hjelm2018learning,isola2015learning,he2022masked,gidaris2018unsupervised,zhang2016colorful,larsson2016learning}, text and its context\cite{kong2019mutual,logeswaran2018efficient,yang2019xlnet,doersch2015unsupervised,zhang2021supporting,su2022contrastive}, different time slots of the same videos \cite{zhuang2020unsupervised,sermanet2018time,miech2020end,sun2019learning,chen2022frame,xu2021rethinking}, which pull similar data pairs close while push dissimilar pairs apart.

The intuition based on the contrastive idea is how to choose similar and dissimilar pairs, one feasible methodology is to maximize the shared information of different views (mutual information). Prior works \cite{oord2018representation,chen2020simple} have shown appealing performance in multiple downstream tasks according to this intuition. However, a few researchers draw some conclusions against intuition. The work by \cite{tschannen2019mutual} argues that maximizing the tighter bound of the mutual information between different variables may lead to worse representations, the success of these promising results should be also attributed to the parametrization of the employed mutual information estimators and the choice of encoder extractor architectures. Therefore, they design several experiments to verify their hypothesis.

The work \cite{tian2020makes} demonstrates that the optimal view for contrastive representation learning is related to the given downstream tasks, meaning no need to maximize the mutual information of different views. Their InfoMin rule aims to figure out particular data augmentations to reduce the mutual information appropriately but does not find what component results in their hypothesis and does not give a general objective function, while our method considers standard augmentations (e.g., cropping, rotation, and colorization), theoretically analyzes the task-irrelevant information between different augmentations and designs a new objective function to eliminate this part. On the other hand, \cite{tsai2020self} reveals that contrastive representation learning is able to extract task-relevant information and discard task-irrelevant information with a fixed gap and quantifies the amount of information that cannot be discarded. They create a new composite self-supervised learning objective based on their analysis, but their introduced Inverse Predictive Learning seems slightly not related to their analysis logic.

What's more, \cite{Federici2020LearningRR} applies information bottleneck \cite{tishby99information} to the multi-view learning, also aims to discard task-irrelevant information from different views during their framework, but their  method is not implemented in a precise way and not tested on frequently-used datasets, our work presents a totally different and flexible objective function, and is validated on popular datasets, such as CIFAR10 \cite{krizhevsky2009learning}, STL-10 \cite{coates2011analysis}, ImageNet \cite{deng2009imagenet}. A recent work \cite{wang2022rethinking} refutes the conclusion given by \cite{tian2020makes,Federici2020LearningRR}, argues that the minimal sufficient representation contains less task-relevant information than other sufficient representations and has a non-ignorable gap with the optimal representation, which may cause performance degradation for several downstream tasks, suggesting increasing the mutual information between the input data and its encoding in the objective function. However, this adjustment maybe can bring more useful information but is also likely to introduce more noise for different downstream tasks. 


\begin{figure*}
\centering
\includegraphics[width=0.8\linewidth]{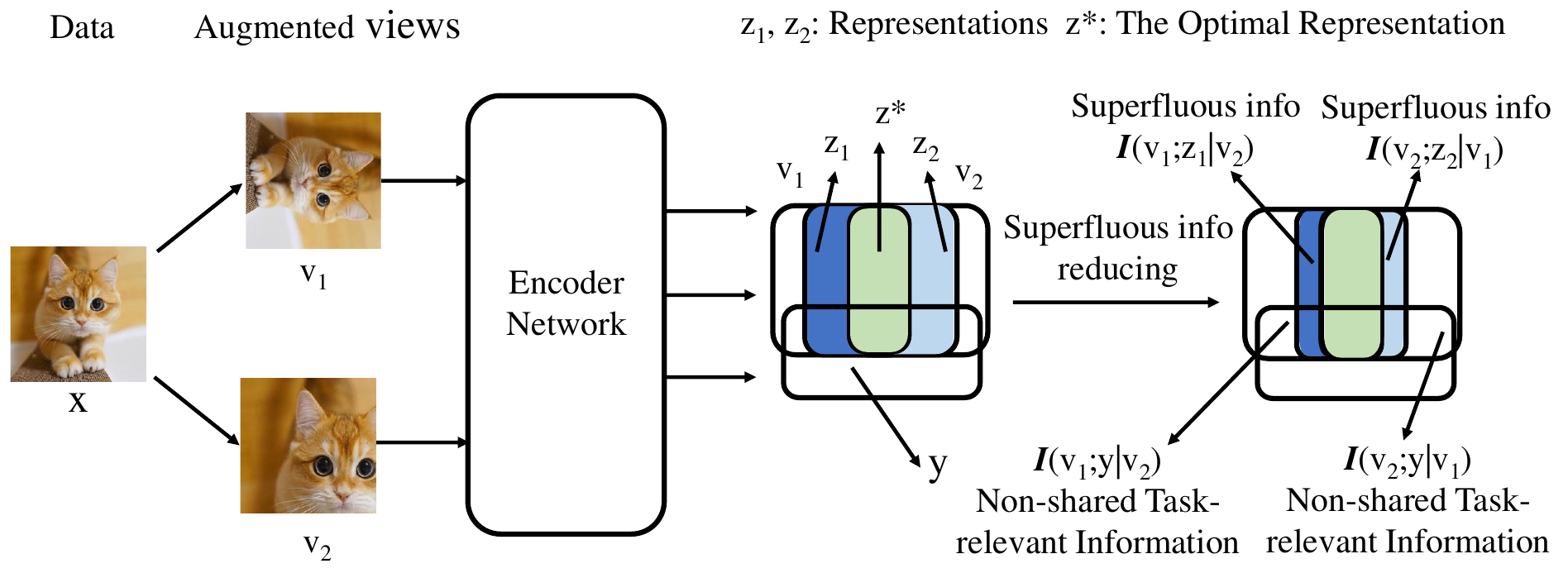}
\caption{The information process of classical contrastive representation learning. We aim to reduce the superfluous information to make the learned representation more sufficient and robust. Meanwhile, the Non-shared task-relevant information sometimes needs to be considered.}
\label{fig2}
\end{figure*}

\section{Method}
\subsection{Motivation}
\label{analysis}
Let us turn to supervised representation learning, the objective of supervised representation learning is to find a good representation $\textbf{z}$ after encoding the input data $\textbf{x}$, and then use the representation $\textbf{z}$ for various downstream tasks, such as classification, regression. Since the label $\textbf{y}$ can be obtained in supervised representation learning, the training metric is usually built up with the representation $\textbf{z}$ and the label $\textbf{y}$. What's more, to make the representation more general and more robust, \cite{alemi2016deep} applies the Information Bottleneck theory \cite{tishby99information} to establish a new objective function, the purpose is to make the representation $\textbf{z}$ more sufficient for the label $\textbf{y}$. We discuss the concept of sufficiency of supervised representation learning by the following definition. 
\begin{definition}
\label{def1}
Sufficiency in supervised representation learning: A representation $\textbf{z}$ of the input data $\textbf{x}$ is \textbf{sufficient} for the label $\textbf{y}$ if and only if $I(\textbf{x};\textbf{y}|\textbf{z})=0$ (Where $I(\cdot)$ represents the mutual information between variables).
\end{definition}

According to Definition 1, we know that the learned sufficient representation $\textbf{z}$ contains all the information related to the label $\textbf{y}$ after the model properly encodes the original input data $\textbf{x}$, and it may be well-performed for different evaluation tasks. Since the input data $\textbf{x}$ usually has high-level semantic information compared to the label $\textbf{y}$, there certainly exists some information in $\textbf{x}$ which is irrelevant for $\textbf{y}$, we can regard these as task-irrelevant (superfluous) information. By decomposing  $I(\textbf{x};\textbf{z})$ into two terms using the chain rule of mutual information (proof in Appendix \ref{b1}).
\begin{align}
    \label{eq1}
    \setlength{\abovedisplayskip}{100pt}
    \setlength{\belowdisplayskip}{3pt}
    I(\textbf{x};\textbf{z}) =  \underbrace{I(\textbf{y};\textbf{z})}_{\text{predictive information}} + \underbrace{I(\textbf{x};\textbf{z}|\textbf{y})}_{\text{superfluous information}}
\end{align}

The conditional mutual information $I(\textbf{x};\textbf{z}|\textbf{y})$ expresses the mutual information between $\textbf{x}$ and $\textbf{z}$, which is task-irrelevant for $\textbf{y}$, so this is superfluous. It is better to make this term as small as possible. While the other term $I(\textbf{y};\textbf{z})$ represents how much task-relevant information contained in the representation, which we want to maximize. Obviously reducing the amount of superfluous information can be done directly in supervised learning. As a consequence, \cite{alemi2016deep} combines two terms $I(\textbf{x};\textbf{z})$ and $I(\textbf{y};\textbf{z})$ to make the model learn a more sufficient representation.

Since there is label information in the supervised setting, we can easily analyze superfluous information and useful information. As for self-supervised representation learning, there are only different augmentation views from the unlabeled data, the only useful information to be leveraged is the shared information between different views. Consider $\textbf{v}_1$ and $\textbf{v}_2$ as two different views of data $\textbf{x}$ and let $\textbf{y}$ be its label. Similarly $\textbf{z}_1$ and $\textbf{z}_2$ become representations of two different views $\textbf{v}_1$ and $\textbf{v}_2$ after processed by the network. Therefore, the main objective is to obtain as much shared information of two views as possible, usually maximizing the mutual information of two representations $\textbf{z}_1$ and $\textbf{z}_2$ ($I(\textbf{z}_1;\textbf{z}_2)$) is what we pay attention to. Nevertheless, like supervised representation learning, there must be some task-irrelevant (superfluous) information contained in the learned representation. Consequently we want to extract task-relevant and discard task-irrelevant information simultaneously. To formalize this we define \textbf{sufficiency} for self-supervised representation learning. 
\begin{definition}
\label{def2}
Sufficiency in self-supervised representation learning: A representation $\textbf{z}_1$ is \textbf{sufficient} of $\textbf{v}_1$ for $\textbf{v}_2$ if and only if $I(\textbf{z}_1;\textbf{v}_2) = I(\textbf{v}_1;\textbf{v}_2)$.
\end{definition}

Intuitively, $\textbf{z}_1$ is sufficient if the amount of information in $\textbf{v}_1$ about $\textbf{v}_2$ is unchanged by the encoding procedure. Symmetrically, $\textbf{z}_2$ is sufficient of $\textbf{v}_2$ for $\textbf{v}_1$ if and only if  $I(\textbf{v}_1;\textbf{z}_2) = I(\textbf{v}_1;\textbf{v}_2)$.
\begin{definition}
\label{def3}
(Minimal sufficiency in self-supervised representation Learning) The sufficient representation $\textbf{z}_1^{min}$ of $\textbf{v}_1$ is minimal if and only if $I(\textbf{z}_1^{min},\textbf{v}_1)\leq I(\textbf{z}_1,\textbf{v}_1)$, $\forall \textbf{z}_1$ that is sufficient.
\end{definition}

From the above definition, we can see that a sufficient representation contains exactly all the shared information between $\textbf{v}_1$ and $\textbf{v}_2$. Therefore, maintaining the representations sufficient and discarding superfluous information between two views and their representations simultaneously is particularly significant. We can show the following equation by factorizing the mutual information between $\textbf{v}_1$ and $\textbf{z}_1$ into two terms, (similarly for $\textbf{v}_2$ and $\textbf{z}_2$):
\begin{align}
\label{eq2}
    I(\textbf{v}_1; \textbf{z}_1) = \underbrace{I(\textbf{v}_2;\textbf{z}_1)}_{\text{predictive information}} + \underbrace{I(\textbf{v}_1;\textbf{z}_1|\textbf{v}_2)}_{\text{superfluous information}}
\end{align}

Similar to Equation \ref{eq1}, $I(\textbf{v}_1; \textbf{z}_1)$ can be decomposed into predictive information component and superfluous information component. Since $I(\textbf{v}_2;\textbf{z}_1)$ expresses the information between one representation and the other view, it makes a contribution to the task-relevant information between two views. On the other hand, $I(\textbf{v}_1;\textbf{z}_1|\textbf{v}_2)$ means the information contained between $\textbf{v}_1$ and $\textbf{z}_1$ while the view $\textbf{v}_2$ has been observed, as shown in Figure \ref{fig2}. The larger this term, the more non-shared information between two views, so reducing (or minimizing) $I(\textbf{v}_1;\textbf{z}_1|\textbf{v}_2)$ can make the learned representation more sufficient. The proof of Equation \ref{eq2} can be found in appendix \ref{b2}.

\subsection{SuperInfo Loss Function}
Since contrastive representation learning tries to pull similar data pairs close while push dissimilar pairs apart, it maximizes the mutual information between two learned representations $\textbf{z}_1$ and $\textbf{z}_2$. Based on the analysis above, it can be concluded that 
reducing the superfluous information may help to learn a more sufficient representation for various downstream tasks, so we can maximize the following objective function.
\begin{align}
\label{}
    J = I(\mathbf{z}_1; \mathbf{z}_2) - \lambda_a I(\mathbf{v}_1;\mathbf{z}_1|\mathbf{v}_2) - \lambda_b I(\mathbf{v}_2;\mathbf{z}_2|\mathbf{v}_1)
\end{align}

Where $J$ represents the objective function, $I(\mathbf{v}_1;\mathbf{z}_1|\mathbf{v}_2)$, $I(\mathbf{v}_2;\mathbf{z}_2|\mathbf{v}_1)$ are the superfluous information of two views analyzed above, ${\lambda}_i (i=a,b)$ are two Lagrangian parameters that we can tune manually.

According to the analysis in section \ref{analysis}, $I(\mathbf{v}_1;\mathbf{z}_1|\mathbf{v}_2) = I(\mathbf{v}_1;\mathbf{z}_1) - I(\mathbf{v}_2; \mathbf{z}_1)$, $I(\mathbf{v}_2;\mathbf{z}_2|\mathbf{v}_1) = I(\mathbf{v}_2;\mathbf{z}_2) - I(\mathbf{v}_1; \mathbf{z}_2)$, and since two augmentation views are symmetric to each other, we set up ${\lambda}_1 = {\lambda}_2$, ${\lambda}_3 = {\lambda}_4$ to make the objective function more general.
\begin{equation}
\label{eq4}
\begin{aligned}      
   J = & I(\mathbf{z}_1; \mathbf{z}_2) - \lambda_1 I(\mathbf{v}_1;\mathbf{z}_1) - \lambda_2 I(\mathbf{v}_2;\mathbf{z}_2) \\
   & + \lambda_3 I(\mathbf{v}_1; \mathbf{z}_2) + \lambda_4 I(\mathbf{v}_2; \mathbf{z}_1)
\end{aligned}
\end{equation}

What's more, since ${\lambda}_i (i=1,2)$ and ${\lambda}_j (i=3,4)$ can be set up differently based on Equation \ref{eq4}, we can adjust these coefficients to discard superfluous information ,while keeping partial non-shared task-relevant information according to different tasks (as shown in Figure \ref{fig2}, non-shared task-relevant information: $I(\mathbf{v}_1;\mathbf{y}|\mathbf{v}_2)$ and $I(\mathbf{v}_2;\mathbf{y}|\mathbf{v}_1)$), this highlights another advantage of our objective function.

We want to maximize the objective function $J$, but it is intractable to deal with mutual information expressions, therefore, we have to maximize the lower bound of $J$ for second best. We first consider the term $I(\mathbf{v}_i;\mathbf{z}_i) (i=1,2)$, 
\begin{equation}
\begin{aligned}
\label{}
    & I(\textbf{v}_i;\textbf{z}_i) = \iint d\textbf{v}_i d\textbf{z}_i p(\textbf{v}_i, \textbf{z}_i) \log \frac{p(\textbf{z}_i | \textbf{v}_i)}{p(\textbf{z}_i)} \\
    & = \iint d\textbf{v}_i d\textbf{z}_i p(\textbf{v}_i, \textbf{z}_i) \log p(\textbf{z}_i | \textbf{v}_i) - \int d\textbf{z}_i p(\textbf{z}_i)  \log p(\textbf{z}_i) 
\end{aligned}
\end{equation}

In general, computing the marginal distribution of $\textbf{z}_i$ might be difficult. Make $r(\textbf{z}_i)$ is a variational approximation to this marginal, since $\textbf{KL}[p(\textbf{z}_i ),r(\textbf{z}_i)] \geq 0$, we can get the following upper bound of $I(\mathbf{v}_i;\mathbf{z}_i)$, in Equation \ref{upper}, we further assume the encoder process follows the Gaussian distribution, $ p(\textbf{z}_i | \textbf{v}_i) = \mathcal{N}(\textbf{z}_i;{f}_i(\textbf{v}_i)),\sigma_i^2 I)$ and the variational approximation $r(\textbf{z}_i) = \mathcal{N}(\textbf{0},I)$, so we can handle the $\textbf{KL}$ divergence terms(full proof in Appendix \ref{MI}).
\begin{equation}
\label{upper}
\begin{aligned}
    I(\textbf{v}_i;\textbf{z}_i) \leq \iint d\textbf{v}_i d\textbf{z}_i p(\textbf{v}_i, \textbf{z}_i) \log \frac{p(\textbf{z}_i | \textbf{v}_i)}{r(\textbf{z}_i)}
\end{aligned}
\end{equation}

On the other hand, we need the lower bound of the positive terms $I(\textbf{v}_1;\textbf{z}_2)$ and $I(\textbf{v}_2;\textbf{z}_1)$, take $I(\textbf{v}_1;\textbf{z}_2)$ as the example. Using the relationship between mutual information expression and entropy expression $I(\mathbf{v}_1;\mathbf{z}_2) = H(\mathbf{v}_1) - H(\mathbf{v}_1|\mathbf{z}_2)$, where $H(\mathbf{v}_1)$ is a constant given the augmentation view, so we only need to maximize $ - H(\mathbf{v}_1|\mathbf{z}_2) = \mathbb{E}_{p(\mathbf{v}_1,\mathbf{z}_2)}[\log p(\textbf{v}_1 | \textbf{z}_2)]$. Assuming $q(\textbf{v}_1 | \textbf{z}_2)$ is the variational approximation to $p(\textbf{v}_1 | \textbf{z}_2)$ in order to deal with the intractability of this conditional distribution, we have the lower bound of $I(\textbf{v}_1;\textbf{z}_2)$, in Equation \ref{lower}. Further we suppose $q(\textbf{v}_1|\textbf{z}_2) = \mathcal{N}(\textbf{v}_1;{h}_1(\textbf{z}_2)),\sigma_3^2 I)$, where $h_1$ maps $\mathbf{z}_2)$ to $\textbf{v}_1$ which we can use an compact \textbf{deConvNet} for realization, thus, we can estimate $\mathbb{E}_{p(\mathbf{v}_1,\mathbf{z}_2)}[\log q(\textbf{v}_1 | \textbf{z}_2)]$. The complete proof can be found in Appendix \ref{cross MI} (similar to $I(\textbf{v}_2;\textbf{z}_1)$).
\begin{equation}
\label{lower}
\begin{aligned}
    I(\textbf{v}_1;\textbf{z}_2) \geq \mathbb{E}_{p(\mathbf{v}_1,\mathbf{z}_2)}[\log p(\textbf{v}_1 | \textbf{z}_2)]
\end{aligned}
\end{equation}

To sum up, we are able to maximize the lower bound of the objective function, so the loss function $L$ is listed in Equation \ref{eq8}, where $I(\mathbf{z}_1; \mathbf{z}_2)$ can be estimated by MINE estimator \cite{belghazi2018mine}, JS divergence estimator \cite{hjelm2018learning}, InfoNCE loss \cite{oord2018representation}. We name our loss ``SuperInfo'' loss, the algorithm is on the right.
\begin{equation}
\label{eq8}
\begin{aligned}
    L  = & - I(\mathbf{z}_1; \mathbf{z}_2) + \sum_{i=1}^2 {\lambda}_i \textbf{KL}[p(\textbf{z}_i | \textbf{v}_i),r(\textbf{z}_i)] \\
    & - {\lambda}_3 \mathbb{E}_{p(\mathbf{v}_1,\mathbf{z}_2)}[\log q(\textbf{v}_1 | \textbf{z}_2)] - {\lambda}_4 \mathbb{E}_{p(\mathbf{v}_2,\mathbf{z}_1)}[\log q(\textbf{v}_2 | \textbf{z}_1)]
\end{aligned}
\end{equation}

\subsection{Bayes Error Rate of Contrastive Learning Representations}
\label{Bayes}

In this subsection, we apply Bayes error rate \cite{feder1994relations} to analyze the irreducible error of self-supervised contrastive learning representations. Suppose the downstream task is classification and $T$ represents the categorical variable. It represents the smallest acceptable error when estimating the correct label given any learned representations. Basically, let $P_e$ be the Bayes error rate of arbitrary self-supervised learning representations $\mathbf{z}_1$ and $\hat{T}$ be the prediction for $T$ from our classification model. According to \cite{feder1994relations}, $P_e=1-\mathbb{E}_{p(\mathbf{z}_1)}[\max_{t \in T} p(\hat{T}=t|z_1)]$, so $0\leq P_e\leq 1-1/|T|$ where $|T|$ is the cardinality of $T$. We define a threshold function $\Gamma(x)=\min\{\max\{x,0\},1-1/|T|\}$ for better analysis. \cite{wang2022rethinking} has proved the following theory (Full proof can be found in this paper).

\begin{theorem}
\cite{wang2022rethinking}
\label{the1}
(Bayes Error Rate of Representations) For arbitrary self-supervised learning representation $\mathbf{z}_1$, its Bayes error rate $P_e=\Gamma(\Bar{P}_e)$ with
\begin{equation}
\label{eq9}
    \Bar{P}_e\leq 1-\exp[-(H(T)-I(\mathbf{z}_1,T|\mathbf{v}_2)-I(\mathbf{z}_1,\mathbf{v}_2,T))]
\end{equation}
Thus, when the learned representation $\mathbf{z}_1^{suf}$ is sufficient, its Bayes error rate $P_e^{suf}=\Gamma(\Bar{P}_e^{suf})$ with
\begin{equation}
\label{eq10}
    \resizebox{.85\linewidth}{!}{$\Bar{P}_e^{suf}\leq1-\exp[-(H(T)-I(\mathbf{z}_1^{suf},T|\mathbf{v}_2)-I(\mathbf{v}_1,\mathbf{v}_2,T))]$}
\end{equation}
Further for the minimal sufficient representation $\mathbf{z}_1^{min}$, its Bayes error rate $P_e^{min}=\Gamma(\Bar{P}_e^{min})$ with
\begin{equation}
\label{eq11}
    \Bar{P}_e^{min}\leq 1-\exp[-(H(T)-I(\mathbf{v}_1,\mathbf{v}_2,T))]
\end{equation}
\end{theorem}

Observing Equation \ref{eq10} and \ref{eq11}, we have that $P_e^{min}$ has a larger upper bound than $P_e^{suf}$ since $I(\mathbf{z}_1^{suf},T|\mathbf{v}_2) \geq 0$. Therefore, \cite{wang2022rethinking} argues increase $I(\textbf{v}_1;\textbf{z}_1)$ to introduce more information that is relevant to different downstream tasks, but it also brings certain "noise", while we can adjust the coefficients of Equation \ref{eq4} to keep partial non-shared task-relevant information according to different tasks. This improvement provides us with a trade-off between the sufficiency of the learned representations and its Bayes error rate.

\begin{algorithm}[!t]
\caption{\label{alg:main} Training with SuperInfo}
\begin{algorithmic}
    \State \textbf{input:} batch size $N$, temperature $\tau$, hyperparameter ${\lambda}_i (i=1,2,3,4)$, encoder network $f$, $g$, data transformation $\mathcal{T}$, sampling network $q_{\mu}$ (MLP), $q_{\sigma}$ (MLP), reconstruct network $r$ (deConvNet).
    \For{sampled minibatch $\{\bm{x_k} \}_{k=1}^N$}
        \State \textbf{for all} $k\in \{1, \ldots, N\}$ \textbf{do}
            \State $~~~~$$t \!\sim\! \mathcal{T}$, $t' \!\sim\! \mathcal{T}$
            \State $~~~~${// two augmentations} 
            \State $~~~~$$\tilde{\bm x}_{2k-1} = t(\bm x_k)$, $\tilde{\bm x}_{2k} = t'(\bm x_k)$
            \State $~~~~${// networks process to get the representation}
            \State $~~~~$$\bm h_{2k-1} = f(\tilde{\bm x}_{2k-1})$, $\bm h_{2k} = f(\tilde{\bm x}_{2k})$
            \State $~~~~${// obtain $\mu$ and $\sigma$ of assumed Gaussian distribution}
            \State $~~~~$$\bm \mu_{2k-1} = q_{\mu}({\bm h}_{2k-1})$, $\bm \mu_{2k} = q_{\mu}({\bm h}_{2k})$ 
            \State $~~~~$$\bm \sigma_{2k-1} = q_{\sigma}({\bm h}_{2k-1})$, $\bm \sigma_{2k} = q_{\sigma}({\bm h}_{2k})$

            \State $~~~~${// reconstruct input from the representations}
            \State $~~~~$$\bm x_{2k-1}^{'} = r({\bm h}_{2k-1})$, $\bm x_{2k}^{'} = r({\bm h}_{2k})$
            
            \State $~~~~${// get the projection after the projection head }
            \State $~~~~$$\bm z_{2k-1} = g({\bm h}_{2k-1})$, $\bm z_{2k} = g({\bm h}_{2k})$
        \State \textbf{end for}
        \State \textbf{for all} $i\in\{1, \ldots, 2N\}$ and $j\in\{1, \dots, 2N\}$ \textbf{do}
        \State $~~~~$ {1) apply method to estimate $- I(\bm z_i; \bm z_j)$}
        \State $~~~~$ {2) calculate Gaussian distribution \textbf{KL} divergence}
        \State $~~~~$ {using $\bm \mu_i$, $\bm \mu_j$, $\bm \sigma_i$, $\bm \sigma_j$}
        \State $~~~~$ {3) calculate distance $\textbf{Dis}(\bm x_i, \bm x_i^{'})$ and $\textbf{Dis}(\bm x_j, \bm x_j^{'})$}
        \State \textbf{end for}
        \State $L_{CL} = \frac{1}{2N} \sum\limits_{i,j=1}^{2N} (- I(\bm z_i; \bm z_j))$
        \State $L_{KL} = \frac{1}{2N} \sum\limits_{i,j=1}^{2N} (\lambda_1 KL(\bm \mu_i,\bm \sigma_i) + \lambda_2 KL(\bm \mu_j,\bm \sigma_j))$
        \State $L_{RE} = \frac{1}{2N} \sum\limits_{i,j=1}^{2N} (-\lambda_3 \textbf{Dis}(\bm x_i, \bm x_i^{'}) - \lambda_4 \textbf{Dis}(\bm x_j, \bm x_j^{'}))$
        \State $\mathcal{L} = L_{CL} + L_{KL} + L_{RE}$
        \State update networks $f$ and $g$ to minimize $\mathcal{L}$
    \EndFor
    \State \textbf{output} encoder network $f(\cdot)$
\end{algorithmic}
\end{algorithm}

\begin{table*}[t]
\centering
\caption{Linear evaluation accuracy ($\%$) from CIFAR10 and STL-10 with the standard ResNet-18 backbone network.}
\scalebox{0.8}
{
    \begin{tabular}{l|c|cccccc}
    \toprule[1.5pt]
    Method      & CIFAR10 & DTD & MNIST & FaMNIST  & VGGFlower & CUBirds & TrafficSigns \\ 
    \midrule[0.5pt]
    CMC \cite{tian2020contrastive}    & 85.06 & 28.77 & 96.48 & 87.91 & 41.67  & 8.19  & 91.62 \\
    BYOL \cite{grill2020bootstrap}    & 85.64 & 31.22 & 97.15 & 88.92 & 40.90  & 8.84  & 92.17 \\
    SimCLR \cite{chen2020simple}      & 85.70 & 29.52 & 97.03 & 88.36  & 42.81  & 8.87 & 92.41 \\
    MIB \cite{Federici2020LearningRR} & 85.68 & 32.66 & 97.57  & 89.31 & 44.79 & 8.95 & 93.36 \\
    SSL Composite \cite{tsai2020self} & 85.90 & 33.25 & 97.72  & 89.72 & 51.82 & 9.88 & 94.58 \\ 
    InfoCL+RC \cite{wang2022rethinking}  & 85.78 & 33.67 & 97.99 & 90.31 & \textbf{54.16} & 10.89 & 95.84 \\
    InfoCL+LBE \cite{wang2022rethinking} & 85.45 & 34.52 & 97.94 & 89.26 & 54.10  & 10.60 & 94.96 \\
    SuperInfo(ours)  & \textbf{86.38} & \textbf{34.86} & \textbf{98.11} & \textbf{91.42} & 53.79 & \textbf{12.16} & \textbf{96.06} \\   
    \toprule[1.5pt]
    Method      & STL-10 & DTD & MNIST  & FaMNIST & VGGFlower & CUBirds & TrafficSigns \\
    \midrule[0.5pt]
    CMC \cite{tian2020contrastive}    & 78.03 & 37.99 & 94.07 & 86.92 & 48.71 & 7.52 & 75.89 \\
    BYOL \cite{grill2020bootstrap}    & 80.83 & 40.05 & 94.45 & 87.23 & 49.41 & 8.54 & 77.54 \\
    SimCLR \cite{chen2020simple}      & 78.86 & 39.41 & 95.00 & 87.31 & 49.41 & 8.34 & 80.25 \\
    MIB \cite{Federici2020LearningRR} & 79.09 & 40.91 & 96.78 & 88.47 & 52.65 & 9.88 & 85.48 \\
    SSL Composite \cite{tsai2020self} & 79.56 & 42.88 & 97.04 & 89.82 & 57.61 & 10.86 & 94.56 \\ 
    InfoCL+RC \cite{wang2022rethinking}  & 79.21 & 41.81 & 97.48 & 89.98 & \textbf{60.46} & 10.03 & \textbf{94.73} \\
    InfoCL+LBE \cite{wang2022rethinking} & 80.17 & 42.07 & 97.04 & 88.68 & 58.51 & 10.11 & 87.77 \\
    SuperInfo(ours)  & \textbf{82.24} & \textbf{44.15} & \textbf{97.85}  & \textbf{90.69} & 57.93 & \textbf{12.87} & 94.64 \\ 
    \bottomrule[1.5pt]
    \end{tabular}
}
\label{tab1}
\end{table*}


\begin{table*}[t]
\centering
\caption{Linear evaluation accuracy ($\%$) from ImageNet with the standard ResNet-50 backbone network.}
\scalebox{0.8}
{
    \begin{tabular}{l|c|cccccc}
    \toprule[1.5pt]
    Method      & ImageNet  & CIFAR10 & CIFAR100 & DTD & VGGFlower & CUBirds & TrafficSigns \\ 
    \midrule[0.5pt]
    CMC \cite{tian2020contrastive}    & 58.87 & 80.96 & 58.61 & 68.96 & 92.85 & 35.26 & 95.03 \\
    BYOL \cite{grill2020bootstrap}    & 61.55 & 82.95 & 61.65 & 70.86 & 94.08 & 36.97 & 95.92 \\
    SimCLR \cite{chen2020simple}      & 61.01 & 82.30 & 59.86 & 70.16 & 93.52 & 36.49 & 95.27 \\
    MIB \cite{Federici2020LearningRR} & 61.11 & 82.68 & 60.79 & 70.91 & 93.66 & 37.09 & 96.07 \\
    SSL Composite \cite{tsai2020self} & 61.62 & 82.89 & 61.97 & 71.08 & \textbf{95.08} & 37.71 & 96.61 \\
    InfoCL+RC \cite{wang2022rethinking}  & 61.60 & 83.30 & 63.56 & 71.22 & 94.53 & 37.42 & 96.47 \\
    InfoCL+LBE \cite{wang2022rethinking} & 61.37 & 83.20 & 61.99 & 70.95 & 94.34 & 37.78 & 95.99 \\
    SuperInfo(ours)  & \textbf{62.24} & \textbf{83.89} & \textbf{64.08} & \textbf{72.37} & 94.96 & \textbf{39.36} & \textbf{96.90} \\ 
    \bottomrule[1.5pt]
    \end{tabular}
}
\label{tab2}
\end{table*}

\section{Experiments}
In this section, we verify our new SuperInfo loss through several experiments. Based on our experimental results, we also provide specific analysis.

\subsection{Verifying the Role of Superfluous Information}
\label{4.1}
We apply the SuperInfo loss to classical contrastive representation learning framework, and pre-train the model on CIFAR10 \cite{krizhevsky2009learning}, STL-10 \cite{coates2011analysis}, and ImageNet \cite{deng2009imagenet}, the learned representation is used for different downstream tasks:  classification, detection and segmentation. We choose previous work as the baselines: CMC \cite{tian2020contrastive}, SimCLR \cite{chen2020simple}, BYOL \cite{grill2020bootstrap}, MIB \cite{Federici2020LearningRR}, Composite SSL \cite{tsai2020self}, InfoCL \cite{wang2022rethinking} (There are several baselines that are not tested on the three datasets, we try our best to get the results).

\bfsection{Data augmentations.} We use the similar set of image augmentations as in SimCLR \cite{chen2020simple}. For CIFAR10 \cite{krizhevsky2009learning} and STL-10 \cite{coates2011analysis}, random cropping, flip and random color distortion are applied, and for ImageNet \cite{deng2009imagenet}, a random patch of the image is selected and resized to 224 × 224 with a random horizontal flip, followed by a color distortion, consisting of a random sequence of brightness, contrast, saturation, hue adjustments, and an optional grayscale conversion. Finally Gaussian blur and solarization are applied to the patches.

\bfsection{Architecture.} We train ResNet-18 \cite{he2016deep} for CIFAR10 \cite{krizhevsky2009learning}, STL-10 \cite{coates2011analysis} whose output is a 512-dim vector, then we apply an MLP to get a 128-dim vector that can be used for $I(\mathbf{z}_1;\mathbf{z}_2)$ estimation. For ImageNet \cite{deng2009imagenet}, we use ResNet-50 \cite{he2016deep} whose output is a 2048-dim vector, then we apply an MLP to get the projector. The output of the ResNet is used as the representation for downstream tasks. 

\bfsection{Pretrain.} We apply the Adam optimizer \cite{kingma2014adam} with the learning rate 3e-4 to train the ResNet-18 \cite{he2016deep} backbone on CIFAR10 \cite{krizhevsky2009learning} and STL-10 \cite{coates2011analysis} with batch size 256 for 200 epochs, we set ${\lambda}_1 = {\lambda}_2=0.01$, ${\lambda}_3 = {\lambda}_4=0.1$. For ImageNet \cite{deng2009imagenet}, we use the LARS optimizer \cite{you2017scaling} to train the ResNet-50 \cite{he2016deep} backbone with batch size 1024 for 200 epochs, we set the base learning rate to 0.3, scaled linearly with the batch size (LearningRate = 0.3 $\times$ BatchSize/256). In addition, we use a global weight decay parameter of 1.5 $\times$ ${10}^{-6}$ while excluding the biases and batch normalization parameters from both LARS adaptation and weight decay, we set ${\lambda}_1 = {\lambda}_2=0.01$, ${\lambda}_3 = {\lambda}_4=0.1$. While estimating the term $I(\mathbf{z}_1;\mathbf{z}_2)$, we choose InfoNce method.

\begin{table*}
\centering
\scalebox{0.85}
{
\begin{tabular}{l|c|cccccc}
\toprule[1.5pt]
Method      & CIFAR10 & DTD & MNIST & FaMNIST  & VGGFlower & CUBirds & TrafficSigns \\ 
\midrule[0.5pt] 
SuperInfo & 86.38 & \textbf{34.86} & \textbf{98.11} & \textbf{91.42} & \textbf{53.79} & \textbf{12.16} & \textbf{96.06} \\
SuperInfo(${\lambda}_1={\lambda}_2=0$, ${\lambda}_3={\lambda}_4=0.1$) & \textbf{86.51} & 34.47 & 97.99 & 90.77 & 53.19 & 11.88 & 95.64 \\
SuperInfo(${\lambda}_1 = {\lambda}_2=0.01$, ${\lambda}_3 = {\lambda}_4=0$) & 86.29 & 32.69 & 97.22 & 89.09 & 50.17 & 9.89 & 94.13 \\
\toprule[1.5pt]
Method      & STL-10 & DTD & MNIST  & FaMNIST & VGGFlower & CUBirds & TrafficSigns \\
\midrule[0.5pt]
SuperInfo  & 82.24 & \textbf{44.15} & \textbf{97.85} & \textbf{90.69} & \textbf{57.93} & \textbf{12.87} & \textbf{94.64} \\
SuperInfo(${\lambda}_1={\lambda}_2=0$, ${\lambda}_3={\lambda}_4=0.1$) & \textbf{82.43} & 43.16 & 97.46 & 89.87 & 57.16 & 12.61 & 94.19 \\
SuperInfo(${\lambda}_1={\lambda}_2=0.01$, ${\lambda}_3={\lambda}_4=0$) & 82.11 & 42.19 & 97.18 & 88.64 & 56.91 & 10.01 & 93.86 \\
\bottomrule[1.5pt]
\end{tabular}
}
\caption{Linear evaluation accuracy with different loss terms($\%$) from CIFAR10 and STL-10 with the standard ResNet-18 backbone (the best result in bold)}
\label{tab4}
\vspace{-2mm}
\end{table*}

\bfsection{Evaluation.} We first evaluate the learned representation from CIFAR10 \cite{krizhevsky2009learning}, STL-10 \cite{coates2011analysis}, ImageNet \cite{deng2009imagenet} by training a linear classifier on top of the frozen backbone, following the procedure described in \cite{chen2020simple,he2020momentum,tian2020contrastive,grill2020bootstrap}. The linear classifier is comprised of a fully-connected layer followed by softmax trained with the SGD optimizer for 100 epochs. The linear evaluation is conducted on other classification datasets: DTD \cite{cimpoi2014describing}, MNIST \cite{lecun1998gradient}, FashionMNIST \cite{xiao2017fashion}, CUBirds \cite{wah2011caltech}, VGGFlower \cite{nilsback2008automated}, Traffic Signs \cite{houben2013detection} and CIFAR100 \cite{krizhevsky2009learning}, performance is reported using standard metrics for each benchmark. We report the results in Table \ref{tab1} and Table \ref{tab2}.

We can see that SuperInfo beats all previous methods on CIFAR10, STL and ImageNet, improving the state-of-the-art results by $\sim$1\% to 2\%, what's more, the downstream classification results show that SuperInfo outperforms other methods on 6 of the 8 benchmarks, providing only slightly worse performance on VGGFlower and TrafficSigns compared to InfoCL method.

\bfsection{Other vision tasks.} We evaluate our representation on different tasks, object detection and instance segmentation. With this evaluation, we know whether SuperInfo’s representation generalizes beyond classification tasks.

\bfsection{PASCAL VOC object detection \cite{everingham2009pascal}.} The model is Faster R-CNN \cite{ren2015faster} with a backbone of R50-C4 \cite{he2017mask} with BN tuned. We fine-tune all methods end-to-end, The image scale is [480, 800] pixels during training and 800 at inference. The same setup is used for all methods, We evaluate the default VOC metric of AP$_\text{50}$ (i.e., IoU threshold is 50\%) and the more stringent metrics of COCO-style AP and AP$_\text{75}$. Evaluation is on the VOC \texttt{test2007} set. Table \ref{voc_dectect} shows the results fine-tuned on \texttt{trainval2007} ($\sim$16.5k images). SuperInfo is better than all previous counterparts: up to \textbf{+0.9} AP$_\text{50}$, \textbf{+1.6} AP, and \textbf{+2.2} AP$_\text{75}$.

\begin{table}
\footnotesize
\centering
\begin{tabularx}{0.45\textwidth}{l|c|c|c}
    Model & AP & AP$_{50}$ & AP$_{75}$ \\
    \toprule[1pt]
    random initialization & 34.8 & 63.1 & 35.2 \\
    \hline
    CMC                & 45.1 & 75.9 & 47.1 \\
    BYOL               & 47.1 & 77.5 & 48.9 \\
    SimCLR             & 45.5 & 76.2 & 47.5 \\
    MIB                & 46.6 & 77.1 & 48.5 \\
    SSL Composite      & 47.5 & 77.8 & 49.9 \\
    InfoCL+RC          & 48.1 & 78.0 & 50.9 \\
    InfoCL+LBE         & 47.4 & 77.8 & 49.7 \\
    SuperInfo          & 49.7 {\color{green}(+1.6)} & 79.1 {\color{green}(+1.1)} & 53.1 {\color{green}(+2.2)}  \\
\end{tabularx}
\caption{\textbf{Comparison with previous methods on object detection on PASCAL VOC}, fine-tuned on \texttt{trainval2007} and evaluated on \texttt{test2007}. In the brackets are the gaps to the previous best results.}
\label{voc_dectect}
\end{table}

\bfsection{COCO object detection and segmentation \cite{lin2014microsoft}.} The detector is Mask R-CNN \cite{he2017mask} with the R50-C4 backbone \cite{he2017mask}, with BN tuned. The image scale is in [640, 800] pixels during training and is 800 at inference. We fine-tune all methods on the \texttt{train2017} set ($\sim$118k images) and evaluate on \texttt{val2017}, following the default 2× schedule. We report the results in Table \ref{COCO}. According the results, we can see that SuperInfo achieves the state-of-the-art results based on the settings above.

\begin{table}
\footnotesize
\centering
\subfloat[Object detection on COCO]{\label{coco_seg}
\begin{tabularx}{0.45\textwidth}{l|c|c|c}
Model & AP$^{bb}$ & AP$_{50}^{bb}$ & AP$_{75}^{bb}$ \\
\toprule[0.1em]
random initialization & 35.6 & 54.6 & 38.2 \\
\hline
CMC            & 38.0 & 58.0 & 41.7 \\
BYOL           & 39.0 & 58.8 & 42.2 \\
SimCLR         & 38.6 & 58.5 & 41.8 \\
MIB            & 38.9 & 58.7 & 42.2 \\
SSL Composite  & 39.1 & 58.8 & 42.3 \\
InfoCL+RC      & 39.3 & 59.0 & 42.6 \\
InfoCL+LBE     & 39.0 & 58.7 & 42.3 \\
SuperInfo      & 39.9 {\color{green}(+0.6)} & 59.6 {\color{green}(+0.6)} & 43.4 {\color{green}(+0.8)} \\
\end{tabularx}
}

\subfloat[Instance segmentation on COCO]{\label{coco_seg}    
\begin{tabularx}{0.45\textwidth}{l|c|c|c}
Model & AP$^{mk}$ & AP$_{50}^{mk}$ & AP$_{75}^{mk}$ \\
\toprule[0.1em]
random initialization & 31.4 & 51.5 & 33.5 \\
\hline
CMC            & 33.1 & 54.8 & 35.0 \\
BYOL           & 34.1 & 55.4 & 36.3 \\
SimCLR         & 33.9 & 55.2 & 36.0 \\
MIB            & 34.1 & 55.3 & 36.3 \\
SSL Composite  & 34.3 & 55.5 & 36.5 \\
InfoCL+RC      & 34.5 & 55.7 & 36.6 \\
InfoCL+LBE     & 34.2 & 55.4 & 36.4 \\
SuperInfo      & 35.5 {\color{green}(+1.0)} & 56.6 {\color{green}(+0.9)} & 37.8 {\color{green}(+1.2)} \\
\end{tabularx}
}
\caption{\textbf{Object detection and instance segmentation fine-tuned on COCO}: bounding-box AP (AP$^{bb}$) and mask AP (AP$^{mk}$) evaluated on \texttt{val2017}. In the brackets are the gaps to the previous best results. In green are the gaps of at least {\color{green}\textbf{+0.5}} point.}
\label{COCO}
\end{table}


\begin{figure}[hbt]
    \centering
    \subcaptionbox{CIFAR10}{
        \includegraphics[width=0.23\textwidth]{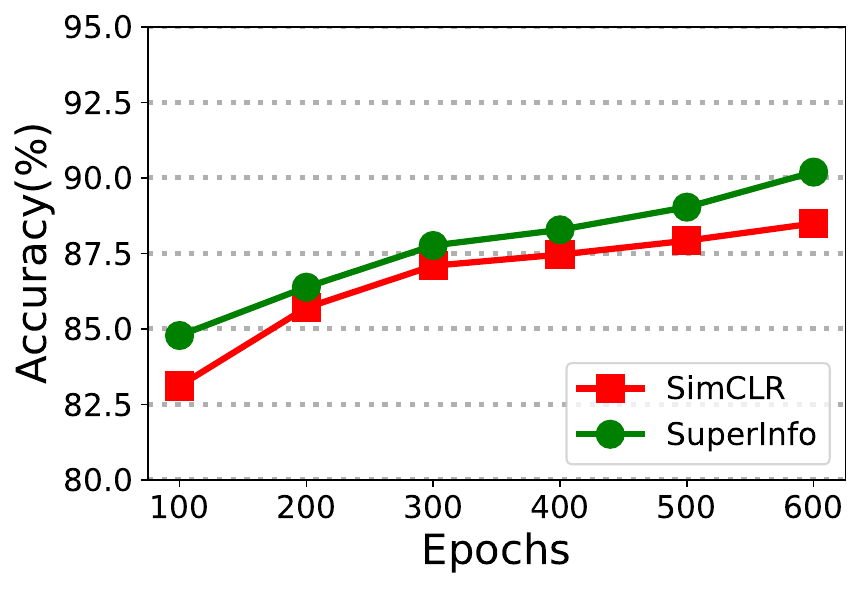}}
    \subcaptionbox{Evaluation from CIFAR10}{
        \includegraphics[width=0.23\textwidth]{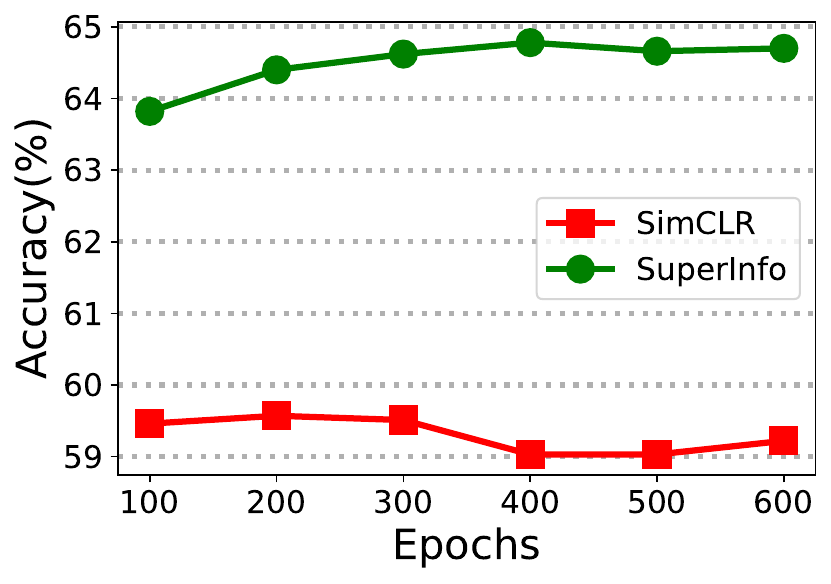}}
    \subcaptionbox{STL-10}{
        \includegraphics[width=0.23\textwidth]{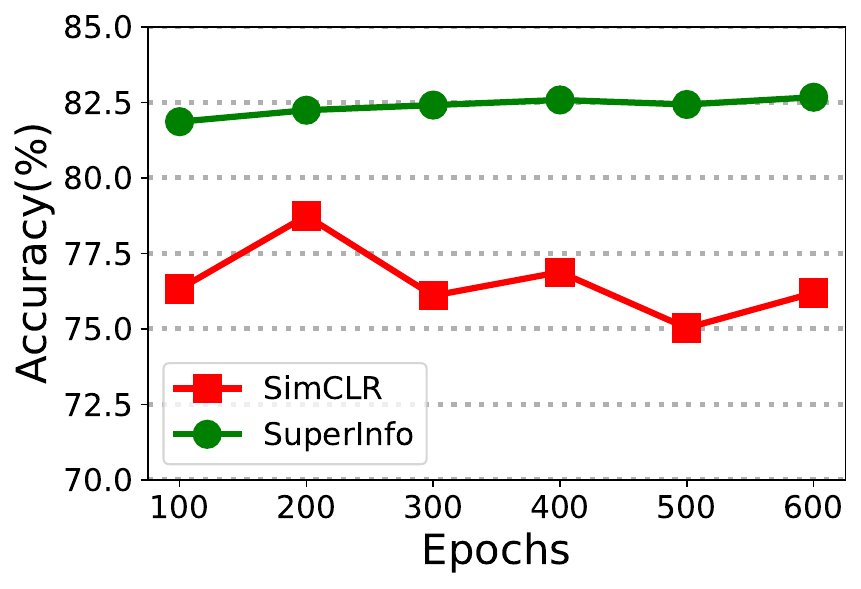}}
    \subcaptionbox{Evaluation from STL-10}{
        \includegraphics[width=0.23\textwidth]{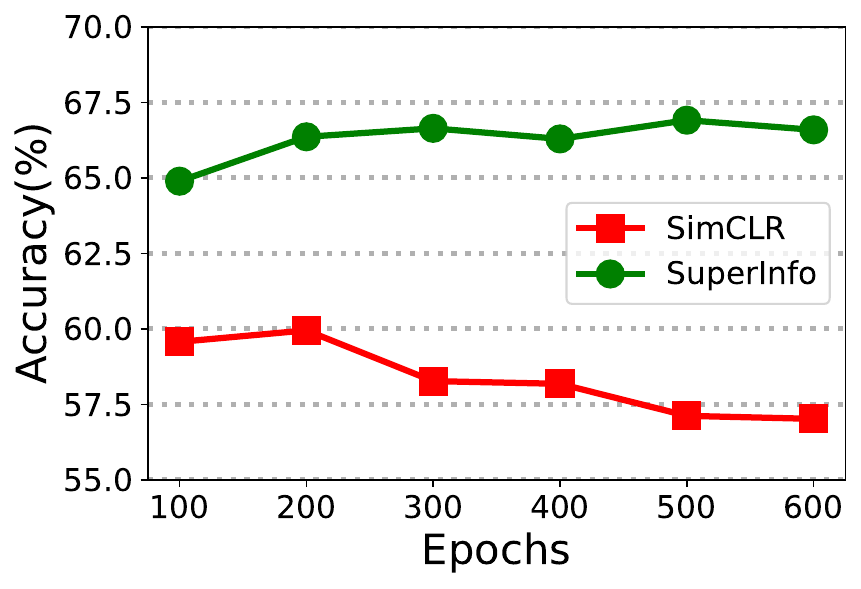}}
    \caption{Classification evaluation accuracy on CIFAR10 and STL-10, and other transfer datasets (Average Accuracy) with training epochs.}
    \label{epoch}
\end{figure}

\subsection{Ablation Study}
\bfsection{Analyzing the role of the added loss terms.} We introduce two new losses into the classical contrastive representation learning loss to make the learned representation more robust and sufficient. Further by analyzing the information flow in the framework (Figure \ref{fig2}), we can adjust the coefficients ${\lambda}_i$ to discard superfluous information, while keeping partial non-shared task-relevant information according to different tasks since the term ($I(\mathbf{v}_1; \mathbf{z}_2), I(\mathbf{v}_2; \mathbf{z}_1)$) introduces certain information from different views which may make contribution to some downstream tasks. Therefore, we conduct the classification experiments (including downstream classification) of only adding the terms, $I(\mathbf{v}_1; \mathbf{z}_1), I(\mathbf{v}_2; \mathbf{z}_2)$ (${\lambda}_3={\lambda}_4=0$) or only adding another terms, $I(\mathbf{v}_1; \mathbf{z}_2), I(\mathbf{v}_2; \mathbf{z}_1)$ (${\lambda}_1={\lambda}_2=0$) to see whether there are apparent different performances with the original SuperInfo (${\lambda}_1={\lambda}_2=0.01$, ${\lambda}_3={\lambda}_4=0.1$). Following the same setting in section \ref{4.1}, we just train the model on CIFAR10 and STL-10 by changing ${\lambda}_i$ and apply linear evaluation protocol to other classification datasets. We report the results in Table \ref{tab4}. It can be clearly seen that the accuracy on the source dataset (CIFAR10 and STL-10) can achieve the similar level compared to original SuperInfo while only adding $I(\mathbf{v}_1; \mathbf{z}_1)$ and $I(\mathbf{v}_2; \mathbf{z}_2)$ since this change can discard superfluous information, but the downstream classification performance gets worse to a certain extent because several non-shared task-relevant information does not keep. On the other hand, only adding $I(\mathbf{v}_1; \mathbf{z}_2)$ and $I(\mathbf{v}_2; \mathbf{z}_1)$ is better than original SuperInfo on the source dataset (CIFAR10 and STL-10), but can not beat original SuperInfo on transfer datasets, which means only introducing non-shared task-relevant information may bring certain noise, leading to the phenomenon of over-fitting on transfer datasets.

\begin{figure}[t]
    \centering
    \subcaptionbox{CIFAR10}{
        \includegraphics[width=0.23\textwidth]{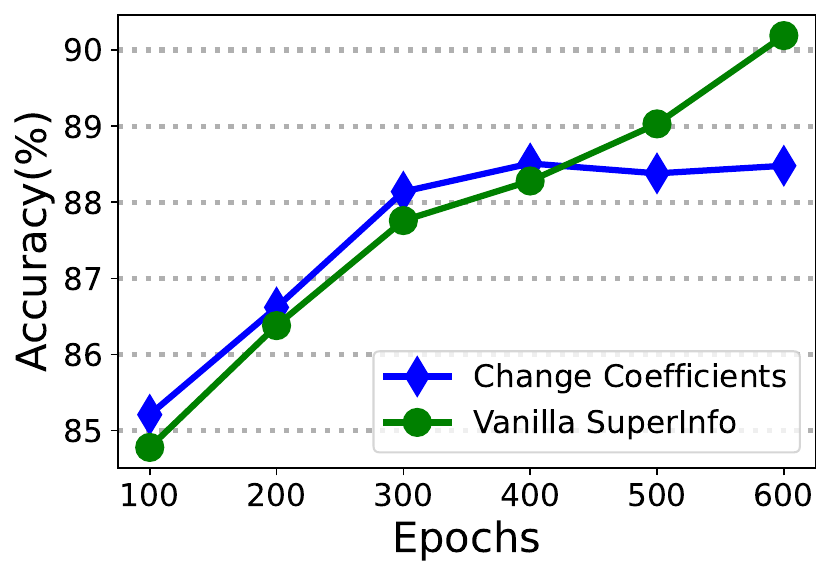}}
    \subcaptionbox{Evaluation from CIFAR10}{
        \includegraphics[width=0.23\textwidth]{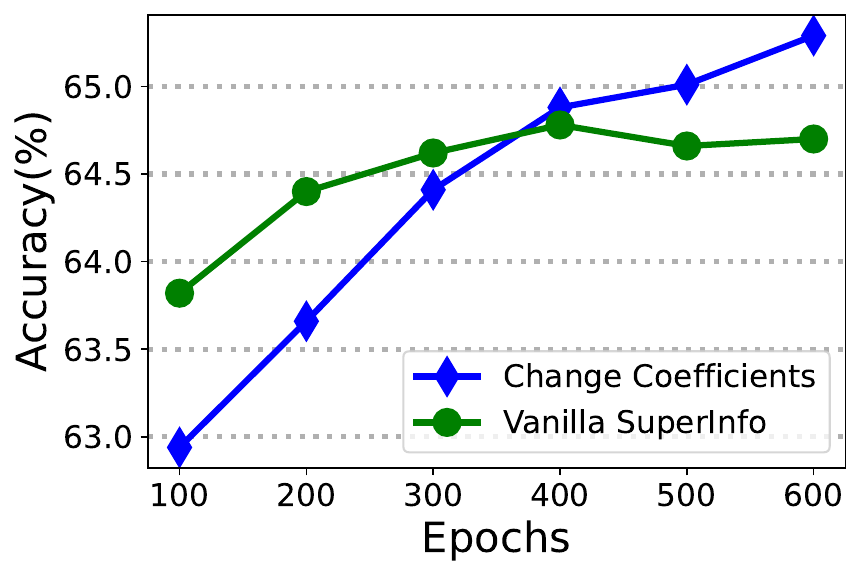}}
    \caption{Classification evaluation accuracy on CIFAR10 and STL-10, and other transfer datasets (Average Accuracy) with training epochs: Vanilla SuperInfo \vs Changing coefficients of SuperInfo}
    \label{change}
\end{figure}

\bfsection{Training with more epochs.} We train all models for 200 epochs during all above experiments. Further we train our model for 100, 200, 300, 400, 500, 600 epochs to analyze SuperInfo's behavior under different training epochs, compared to vanilla SimCLR. The results are listed in Figure \ref{epoch}. According to the above figure, we find that the downstream classification accuracy does not become better with more training epochs, even decreases in the middle period since the learned representations in contrastive representation learning are able to get more close to the minimal sufficient representation which only contains the shared information between different views with more training epochs and the minimal sufficient representation may have the risk of over-fitting on the transfer datasets. This phenomenon is consistent with the conclusion in \cite{wang2022rethinking}. What's more, SuperInfo does bring significant improvements compared to vanilla SimCLR on the transfer datasets under every training epochs. On the other hand, we change the coefficients (${\lambda}_1={\lambda}_2=0.005$, ${\lambda}_3={\lambda}_4=0.5$) compared to vanilla SuperInfo, training the model on CIFAR10 and evaluating the model on other transfer datasets. The results are reported in Figure \ref{change}. As shown in \ref{change}(b), the classification accuracy increases stably with the training epochs, however, the performance on the source dataset (CIFAR10) does not keep the pace with it. This phenomenon shows there really exists a trade-off of the performance on source dataset and on other transfer datasets with respect to the coefficients.



\section{Conclusion and Discussion}
In this work, we deeply analyze the reason why more estimated mutual information between two different views in contrastive representation learning does not guarantee great performance in various downstream tasks and design a new objective function to discards task-irrelevant information, while keeping some non-shared task-relevant information. The effectiveness of our method are verified by several experiments.

There are a few limitations during our presentation. (1) It is still troublesome to determine the coefficients ${\lambda}_i$ of the loss function since they apparently influence the performance, so far we have to tune them manually. (2) Due to our limited computing resources, we can not compare our best results to other methods under the condition of 4096 batch size or larger, and more training epochs, however, a series of better outcomes indicate our method can make contribution to classical contrastive representation learning framework.

{\small
\clearpage
\bibliographystyle{ieee_fullname}
\bibliography{egbib}
}

\clearpage
\appendix

\section{Properties of Mutual Information}
\label{app:mi_properties}
In this section we list some properties \cite{cover1991entropy}
of mutual information and use these properties to prove theorems and equations in this paper. For any random variables $\textbf{x}$, $\textbf{y}$ and $\textbf{z}$, the following equations hold.
\begin{enumerate}
    \item[$(P_1)$] Symmetry:
    \begin{align*}
        I(\textbf{x};\textbf{y}) = I(\textbf{y};\textbf{x})
    \end{align*}
    \item[$(P_2)$] Non-negativity:
    \begin{align*}
        I(\textbf{x};\textbf{y}) \ge 0, I(\textbf{x};\textbf{y}|\textbf{z}) \ge 0
    \end{align*}
    \item[$(P_3)$] Chain rule:
    \begin{align*}
        I(\textbf{x},\textbf{y};\textbf{z}) = I(\textbf{y};\textbf{z})+I(\textbf{x};\textbf{z}|\textbf{y})
    \end{align*}
    \item[$(P_4)$] Multivariate Mutual Information:
    \begin{align*}
        I(\textbf{x};\textbf{y};\textbf{z}) = I(\textbf{y};\textbf{z}) - I(\textbf{y};\textbf{z}|\textbf{x})
    \end{align*}
\end{enumerate}

\section{Proofs}
\label{app:proofs}

Before the detailed derivation, we have the following assumption,if a random variable $\textbf{z}$ is defined to be a representation of another random variable $\textbf{x}$, we state that $\textbf{z}$ is conditionally independent from any other variable in the model once $\textbf{z}$ is observed. This assumption is also reported in \cite{Federici2020LearningRR}.

\begin{align*}
    I(\textbf{z}; \textbf{a}|\textbf{x},\textbf{b}) = 0
\end{align*}

where \textbf{a} and \textbf{b} represent any other variable (or groups of variables) in supervised and self-supervised settings.

\subsection{Proof of Equation \ref{eq1}}
\label{b1}
Equation \ref{eq1}: $I(\textbf{x};\textbf{z})$ can be decomposed into two terms, $I(\textbf{x};\textbf{z}) = I(\textbf{y};\textbf{z}) +  I(\textbf{x};\textbf{z}|\textbf{y})$.

\begin{proof}
        Using Property 3 in Appendix \ref{app:mi_properties}, we see that
        \begin{align*}
            I(\textbf{x},\textbf{y};\textbf{z}) = I(\textbf{y};\textbf{z})+I(\textbf{x};\textbf{z}|\textbf{y})
        \end{align*}
        at the same time, by changing the order of $\textbf{x}$ and $\textbf{y}$, we also have that 
        \begin{align*}
            I(\textbf{x},\textbf{y};\textbf{z}) = I(\textbf{x};\textbf{z})+I(\textbf{y};\textbf{z}|\textbf{x})
        \end{align*}
        Based on the assumption above, $I(\textbf{y};\textbf{z}|\textbf{x}) = 0$, so $I(\textbf{x};\textbf{z}) = I(\textbf{y};\textbf{z}) +  I(\textbf{x};\textbf{z}|\textbf{y})$, Equation \ref{eq1} holds.
\end{proof}

\subsection{Proof of Equation \ref{eq2}}
\label{b2}
Equation \ref{eq2}: $I(\textbf{v}_1; \textbf{z}_1)$ can be decomposed into two terms, $I(\textbf{v}_1; \textbf{z}_1) = I(\textbf{v}_2;\textbf{z}_1) +  I(\textbf{v}_1;\textbf{z}_1|\textbf{v}_2)$.

\begin{proof}
        Using Property 4, we see that
        \begin{align*}
            I(\textbf{v}_1;\textbf{z}_1) = I(\textbf{v}_1;\textbf{z}_1|\textbf{v}_2)+I(\textbf{v}_2;\textbf{v}_1; \textbf{z}_1)
        \end{align*}
        Using Property 4 again, we have that
        \begin{align*}
            I(\textbf{v}_2;\textbf{v}_1; \textbf{z}_1) = I(\textbf{v}_2;\textbf{z}_1) -I(\textbf{v}_2;\textbf{z}_1|\textbf{v}_1)
        \end{align*}
       According to the assumption above, $I(\textbf{v}_2;\textbf{z}_1|\textbf{v}_1) = 0$, so $I(\textbf{v}_1; \textbf{z}_1) = I(\textbf{v}_2;\textbf{z}_1) +  I(\textbf{v}_1;\textbf{z}_1|\textbf{v}_2)$, Equation \ref{eq2} holds.
\end{proof}

\section{Implementation of the objective function}

\subsection{Implementation of $I(\textbf{v}_i;\textbf{z}_i) (i=1,2)$}
\label{MI}
We first expand the $I(\textbf{v}_i;\textbf{z}_i)$ by the definition of mutual information and then apply several operations. Assuming $r(\textbf{z}_i)$ is a variational approximation to this marginal, and because of $p(\textbf{z}_i) = \int d\textbf{v}_i p(\textbf{z}_i|\textbf{v}_i) p(\textbf{v}_i) $ and $\textbf{KL}[p(\textbf{z}_i ),r(\textbf{z}_i)] \geq 0$, we have that

\begin{equation*}
\label{}
\begin{aligned}
    I(\textbf{v}_i;\textbf{z}_i) & = \iint d\textbf{v}_i d\textbf{z}_i p(\textbf{v}_i) p(\textbf{z}_i|\textbf{v}_i) \log \frac{p(\textbf{z}_i|\textbf{v}_i)}{p(\textbf{z}_i)} \\
    & = \iint d\textbf{v}_i d\textbf{z}_i p(\textbf{v}_i) p(\textbf{z}_i|\textbf{v}_i) \log \frac{p(\textbf{z}_i|\textbf{v}_i)}{r(\textbf{z}_i)} \\ 
    & \quad - \iint d\textbf{v}_i d\textbf{z}_i p(\textbf{v}_i) p(\textbf{z}_i|\textbf{v}_i) \log \frac{p(\textbf{z}_i)}{r(\textbf{z}_i)} \\
    & = \iint d\textbf{v}_i d\textbf{z}_i p(\textbf{v}_i,\textbf{z}_i) \log \frac{p(\textbf{z}_i|\textbf{v}_i)}{r(\textbf{z}_i)} \\ 
    & \quad - \int d\textbf{z}_i p(\textbf{z}_i) \log \frac{p(\textbf{z}_i)}{r(\textbf{z}_i)} \\
    & = \iint d\textbf{v}_i d\textbf{z}_i p(\textbf{v}_i,\textbf{z}_i) \log \frac{p(\textbf{z}_i|\textbf{v}_i)}{r(\textbf{z}_i)} \\ 
    & \quad - \textbf{KL}[p(\textbf{z}_i),r(\textbf{z}_i)] \\
    & \leq \iint d\textbf{v}_i d\textbf{z}_i p(\textbf{v}_i,\textbf{z}_i) \log \frac{p(\textbf{z}_i|\textbf{v}_i)}{r(\textbf{z}_i)} \\
    & = \textbf{KL}[p(\textbf{z}_i|\textbf{v}_i),r(\textbf{z}_i)]
\end{aligned}
\end{equation*}

Further we assume the encoder process follows the Gaussian distribution, $ p(\textbf{z}_i | \textbf{v}_i) = \mathcal{N}(\textbf{z}_i;{f}_i(\textbf{v}_i)),\sigma_i^2 I)$ and the variational approximation $r(\textbf{z}_i) = \mathcal{N}(\textbf{0},I)$, so the following equation holds, $d$ is the of embedding vectors. 

\begin{equation*}
\label{}
\begin{aligned}
    \textbf{KL}[p(\textbf{z}_i|\textbf{v}_i),r(\textbf{z}_i)] = -\frac{1}{2} \sum_{i=1}^d(1+\log(\sigma_i^2)-{f}_i^2(\textbf{v}_i)-\sigma_i^2)
\end{aligned}
\end{equation*}

According the above expression, it can be conveniently implemented with code.

\subsection{Implementation of $I(\textbf{v}_1;\textbf{z}_2)$, $I(\textbf{v}_2;\textbf{z}_1)$}
\label{cross MI}
On the other hand, we need the lower bound of the positive terms $I(\textbf{v}_1;\textbf{z}_2)$ and $I(\textbf{v}_2;\textbf{z}_1)$, take $I(\textbf{v}_1;\textbf{z}_2)$ as the example. Assuming $q(\textbf{v}_1 | \textbf{z}_2)$ is the variational approximation to $p(\textbf{v}_1 | \textbf{z}_2)$ in order to deal with the intractability of this conditional distribution, we have the following proof.

\begin{equation*}
\label{}
\begin{aligned}
    I(\textbf{v}_1;\textbf{z}_2) & = \iint d\textbf{v}_1 d\textbf{z}_2 p(\textbf{v}_1,\textbf{z}_2) \log \frac{p(\textbf{v}_1|\textbf{z}_2)}{p(\textbf{v}_1)} \\
    & = \iint d\textbf{v}_1 d\textbf{z}_2 p(\textbf{v}_1,\textbf{z}_2) \log p(\textbf{v}_1|\textbf{z}_2) \\
    & \quad - \iint d\textbf{v}_1 d\textbf{z}_2 p(\textbf{v}_1,\textbf{z}_2) \log p(\textbf{v}_1) \\
    & \approx \iint d\textbf{v}_1 d\textbf{z}_2 p(\textbf{v}_1,\textbf{z}_2) \log q(\textbf{v}_1|\textbf{z}_2) + H(\textbf{v}_1) \\ 
    & = \mathbb{E}_{p(\textbf{v}_1, \textbf{z}_2)}[\log q(\textbf{v}_1| \textbf{z}_2)] + H(\textbf{v}_1)
\end{aligned}
\end{equation*}

Where $H(\mathbf{v}_1)$ is a constant given the augmentation view during objective optimization, so it is equivalent to maximize $ \mathbb{E}_{p(\mathbf{v}_1,\mathbf{z}_2)}[\log p(\textbf{v}_1 | \textbf{z}_2)]$. Further we suppose $q(\textbf{v}_1|\textbf{z}_2) = \mathcal{N}(\textbf{v}_1;{h}_1(\textbf{z}_2)),\sigma_3^2 I)$, where $h_1$ maps $\mathbf{z}_2)$ to $\textbf{v}_1$ which we can use an compact \textbf{deConvNet} for realization, thus, we are able to estimate and implement $\mathbb{E}_{p(\mathbf{v}_1,\mathbf{z}_2)}[\log q(\textbf{v}_1 | \textbf{z}_2)]$. (similar to $I(\textbf{v}_2;\textbf{z}_1)$).

\begin{equation*}
\label{}
\begin{aligned}
    \mathbb{E}_{p(\textbf{v}_1, \textbf{z}_2)}[\log q(\textbf{v}_1| \textbf{z}_2)] \propto -\mathbb{E}_{p(\textbf{v}_1, \textbf{z}_2)}[\|\textbf{v}_1-{h}_1(\textbf{z}_2)\|_2^2] + c 
\end{aligned}
\end{equation*}
where $c$ is a constant to representation $ \textbf{z}_2 $.

Based on these derivations, we can implement the final loss function $L$ 

\begin{equation*}
\label{}
\begin{aligned}
    L  = & - I(\mathbf{z}_1; \mathbf{z}_2) + \sum_{i=1}^2 {\lambda}_i \textbf{KL}[p(\textbf{z}_i | \textbf{v}_i),r(\textbf{z}_i)] \\
    & - {\lambda}_3 \mathbb{E}_{p(\mathbf{v}_1,\mathbf{z}_2)}[\log q(\textbf{v}_1 | \textbf{z}_2)] - {\lambda}_4 \mathbb{E}_{p(\mathbf{v}_2,\mathbf{z}_1)}[\log q(\textbf{v}_2 | \textbf{z}_1)]
\end{aligned}
\end{equation*}

\end{document}